\documentclass[letterpaper,twocolumn,fleqn]{article} 

\usepackage{ist}
\usepackage{cite}
\usepackage{graphicx}
\usepackage[cmex10]{amsmath}
\usepackage{array}
\usepackage{color}
\usepackage{amssymb}
\usepackage{balance}
\usepackage{gensymb}
\usepackage{textcomp}
\usepackage{nameref}

\title{Detection of Object Throwing Behavior in Surveillance Videos}

\author{Ivo P.C. Kersten, Erkut Akdag$^{\dagger}$, Egor Bondarev, and Peter H.N. de With\\
Eindhoven University of Technology, Department of Electrical Engineering, 5612 AP Eindhoven, The Netherlands\\
$^{\dagger}$$\mathit{Corresponding}$ $\mathit{author:}$ $\mathit{e.akdag@tue.nl}$
}

\date{} 

\begin{document} 

\maketitle 

\thispagestyle{empty} 


\begin{abstract}

Anomalous behavior detection is a challenging research area within computer vision. Progress in this area enables automated detection of dangerous behavior using surveillance camera feeds. A dangerous behavior that is often overlooked in other research is the throwing action in traffic flow, which is one of the unique requirements of our Smart City project to enhance public safety. This paper proposes a solution for throwing action detection in surveillance videos using deep learning. At present, datasets for throwing actions are not publicly available. To address the use-case of our Smart City project, we first generate the novel public 'Throwing Action' dataset, consisting of 271 videos of throwing actions performed by traffic participants, such as pedestrians, bicyclists, and car drivers, and 130 normal videos without throwing actions. Second, we compare the performance of different feature extractors for our anomaly detection method on the UCF-Crime and Throwing-Action datasets. The explored feature extractors are the Convolutional 3D (C3D) network, the Inflated 3D ConvNet (I3D) network, and the Multi-Fiber Network (MFNet). Finally, the performance of the anomaly detection algorithm is improved by applying the Adam optimizer instead of Adadelta, and proposing a mean normal loss function  that covers the multitude of normal situations in traffic. Both aspects yield better anomaly detection performance. Besides this, the proposed mean normal loss function also lowers the false alarm rate on the combined dataset. The experimental results reach an area under the ROC curve of 86.10 for the Throwing-Action dataset, and 80.13 on the combined dataset, respectively.

\end{abstract}



\section{Introduction}
\label{Introduction_section}

Surveillance cameras are increasingly deployed to monitor public spaces to enhance safety. The presence of surveillance cameras deters people from misbehaving and accelerates the detection of unsafe behavior in traffic. However, the increasing amount of surveillance footage processed by control-room operators adversely impacts the actual efficiency of these cameras. The human operators try to follow many camera streams at once, which results in a higher likelihood of missing anomaly events. Anomaly events are defined as deviations from the normal patterns in traffic flow. Such events can range from any type of traffic accident up to suspicious behavior of people. Although anomalies in traffic flow occur rarely, they have a significant impact on public safety. Therefore, detecting these anomalies should be automated to ensure maximum efficiency of a surveillance system and improve the detection of anomaly events to sustain public safety. 

The automation of anomaly  detection is a topic that has recently seen an increase in attention from the research community, where a focus is placed on detecting a limited number of severe anomalies. However, certain types of anomalies have received hardly any attention from the research community, even though they can result in dangerous situations. One such anomaly is the action of throwing objects into the surroundings by car drivers or other traffic participants. Such throwing activities can have crucial consequences, from disrupting the traffic flow to creating dangerous situations for all traffic participants. Given the impact of this action, it is worth noting that we have found no research work that aims at the detection of throwing anomalies. For this reason we have constructed a dataset with throwing actions by different traffic participants. Furthermore, this study explores how to apply the anomaly-analysis approaches to the problem of detecting throwing actions and how to merge our throwing action dataset into existing anomaly detection datasets. The ultimate goal of our research is to enable automated detection of dangerous behavior using surveillance camera feeds. A dangerous behavior has been neglected often in other research and contains multiple dimensions, where throwing actions in traffic flow is one of the unique requirements and an important use case in Smart City projects.

\begin{figure}[t!]
    \centering
    \includegraphics[width=0.5\textwidth]{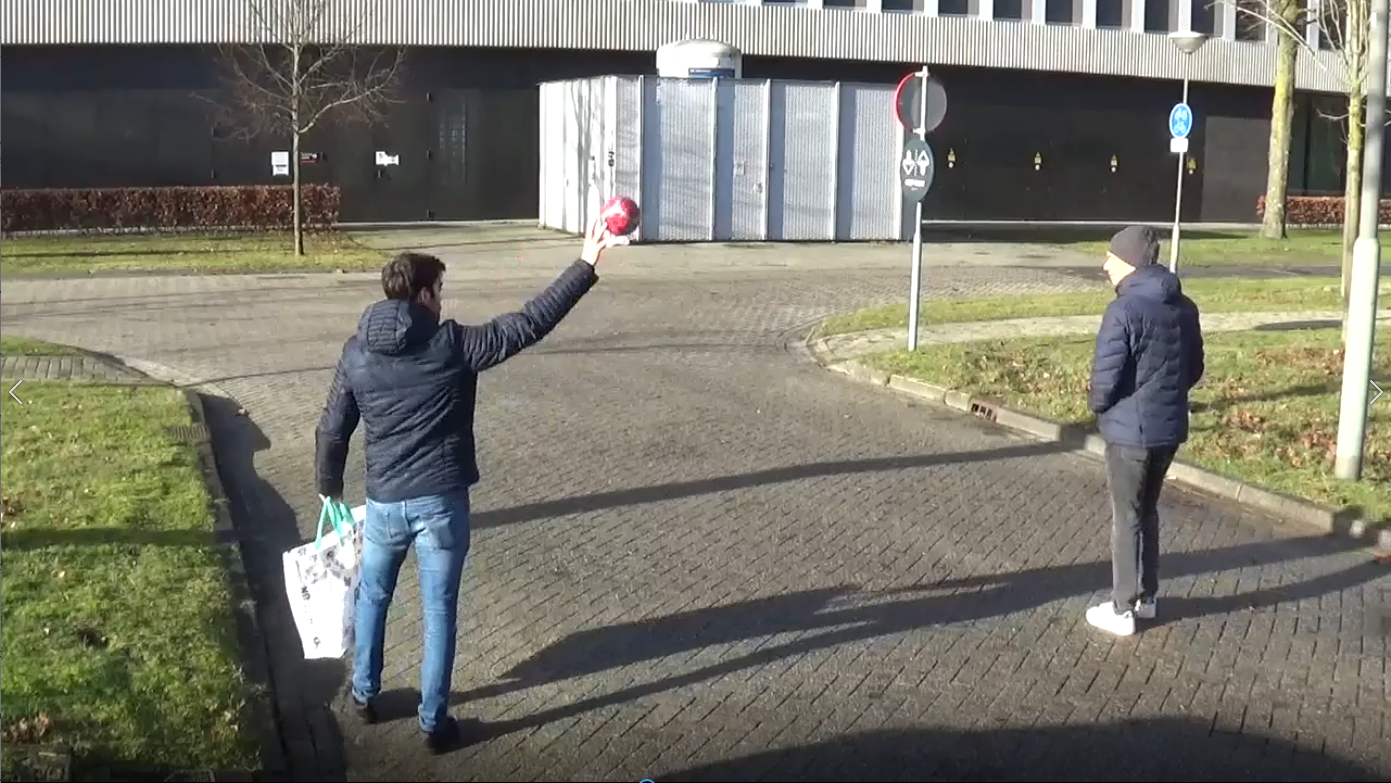}
    \caption{Example frame from one of the videos in the proposed generated Throwing-Action dataset, where a pedestrian is throwing an object at another pedestrian.}
    \label{fig:throwing_action_example}
\end{figure}

This paper is organized as follows. First, the section on ~\nameref{Related_work_section} provides an overview of throwing action detection and general anomaly detection in the literature. Next, in the ~\nameref{Dataset_section} section, we introduce the novel throwing action dataset. The ~\nameref{Methodology_section} section presents our proposed method and contributions. Then we evaluate experimental results in the ~\nameref{Experiment_section}, while in the ~\nameref{Analysis_section} section, the performance of the proposed anomaly detection method is analyzed with respect to the false alarm rate and qualitative results. Finally, the paper is concluded in the ~\nameref{Conclusion_section} section.


\section{Related Work}
\label{Related_work_section}

This section presents the literature overview on methods for throwing action detection and on supervised, unsupervised, and semi-supervised anomaly detection methods.

\subsection{A. Throwing Action Detection}

Related to throwing action detection, previous studies mainly focus on the detection of the thrown object itself instead of the action. In~\cite{Dai2021AIdentification}, a modified version of YOLOv3~\cite{Redmon2018YOLOv3:Improvement} is proposed to detect particular objects when they are thrown out of a car window. This approach relies on the recognition of specific small objects, which is difficult when dealing with low-resolution video footage or throwing actions are performed far away from the camera.

Other studies in~\cite{Csordas2015} and~\cite{Ribnick2007DetectionScenes} concentrate on the trajectories of moving objects. The authors fit parabolic trajectories into the potential trajectories. In other words, any object that is found to follow a parabolic trajectory is deemed as a thrown object. This approach requires long trajectories of thrown objects to reliably fit a parabolic trajectory. In practice, objects are often thrown horizontally or downwards, which results in a short trajectory that renders this method inefficient and only partially addressing the case.

\subsection{B. Anomaly Detection}

Anomaly detection is a challenging computer vision problem aiming to detect rare events from a video stream. Different approaches to this problem can be split into unsupervised, supervised, and semi-supervised categories.

\subsubsection{1) Unsupervised Anomaly Detection} 
The most popular approach is unsupervised anomaly detection, where the training set contains only normal sequences. This approach is employed in many papers, such as~\cite{Zhou2019, Singh2019, Sabokrou2018, Nguyen2019, Chang2020, Deepak2020, Li2020, Nawaratne2020}. During the training stage, a model of normal behavior is constructed. This model can properly characterize normal data in the testing stage, whereas it cannot characterize anomalous data. Applying this method is advantageous when the exact nature of the target anomalies is unknown or uncertain. A drawback of unsupervised anomaly detection is that creating a model capable of exactly capturing all possible normal behaviors is complicated. In other words, similar behaviors can be often both anomalous or normal, depending on the context. For example, a car driving down the right side of the road is considered normal behavior, while driving down the wrong side of the road is not.

\subsubsection{2) Supervised Anomaly Detection}
The second type is supervised training to learn anomalous behavior. Several authors~\cite{Petrocchi2021, Nasaruddin2020, Ullah2020} use supervised anomaly detection methods to predict the presence of anomalies only at the temporal level, while others~\cite{Zhou2019, Liu2019a} also locate the anomalies at the spatial level. In contrast to the unsupervised methods that typically rely on a reconstruction error to find abnormal inputs, supervised learning generally learns to predict an anomaly score directly from the input data. An advantage of the supervised anomaly detection methods is that they often outperform unsupervised techniques for the specific anomalies for which they are trained. One of the drawbacks is the ambiguity of where specifically an anomaly begins and ends. The second drawback is that these methods can only detect the anomalies existing in the dataset, furthermore, annotating the training data is labor-intensive and expensive, as anomaly locations should be specified in every video frame or segment. 

\subsubsection{3) Semi-supervised Anomaly Detection}
The last type of anomaly detection is semi-supervised anomaly detection. This method for anomaly detection is first introduced in~\cite{Sultani2018} and is further expanded in~\cite{Lv2021, Tian2021Weakly-supervisedLearning}. In this approach, the training data contains both videos of normal and anomalous events, however, the data is annotated on a per-video basis. It is significantly faster to annotate data in this way, compared to labeling at individual frames or areas in each frame. This annotation brings a major advantage over the fully-supervised anomaly detection methods. A disadvantage of this approach is that it only learns to recognize anomalies that occur in the training data. 

As can be derived, previous studies mainly focus on object or trajectory detection, instead of considering the throwing action detection as an anomaly detection use case. Therefore, we consider these throwing actions as anomalies, and generate a dataset of throwing actions performed by different road users in real traffic flow. We opted for a semi-supervised anomaly detection method, as opposed to a supervised anomaly detection method, because of its advantages in requiring less labeled data, while still providing a higher accuracy than the unsupervised methods.


\section{Proposed Novel Dataset}
\label{Dataset_section}

\begin{table*}[ht!]
\centering
\begin{tabular}{l|l|c}
\hline
Class                & Definition                                                                                                                       & Examples of video \\ \hline \hline
Bicycle Dangerous    & \begin{tabular}[c]{@{}c@{}}Object is thrown by a bicyclist\\ towards another traffic participant\end{tabular}                    & 
\begin{tabular}[c]{c}\includegraphics[width=0.3\textwidth]{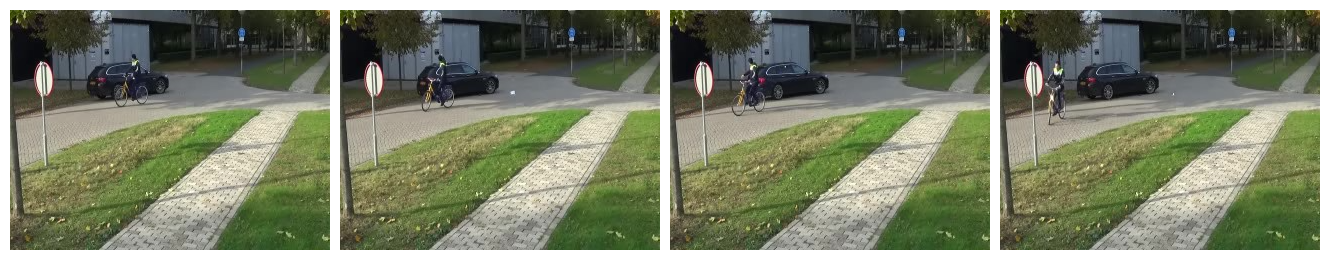}\end{tabular} \\ \hline
Bicycle Safe         & \begin{tabular}[c]{@{}c@{}}Object is throwing by a bicyclist\\ onto the ground\end{tabular}                                      & \begin{tabular}[c]{c}\includegraphics[width=0.3\textwidth]{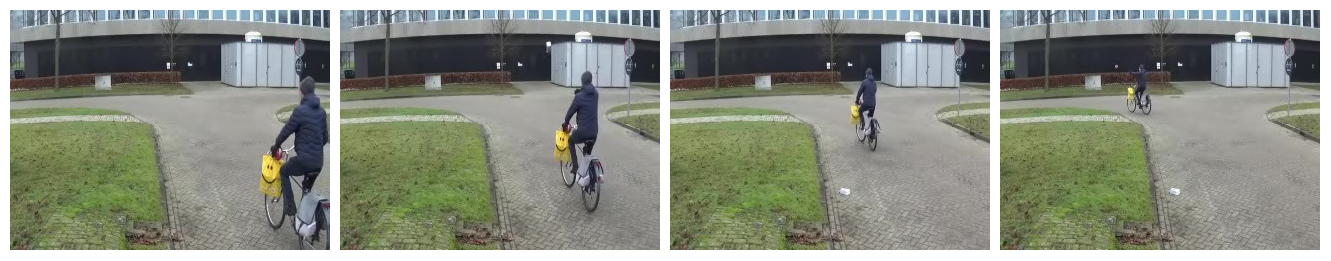}\end{tabular} \\ \hline
Car Dangerous        & \begin{tabular}[c]{@{}c@{}}Object is thrown out of a car window\\ towards another traffic participant\end{tabular}               &
\begin{tabular}[c]{c}\includegraphics[width=0.3\textwidth]{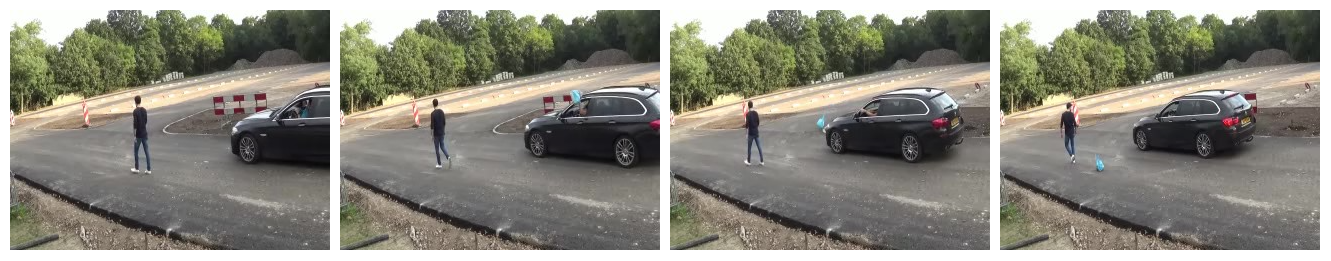}\end{tabular} \\ \hline
Car Safe             & \begin{tabular}[c]{@{}c@{}}Object is thrown out of a car window\\ onto the ground\end{tabular}                                   &
\begin{tabular}[c]{c}\includegraphics[width=0.3\textwidth]{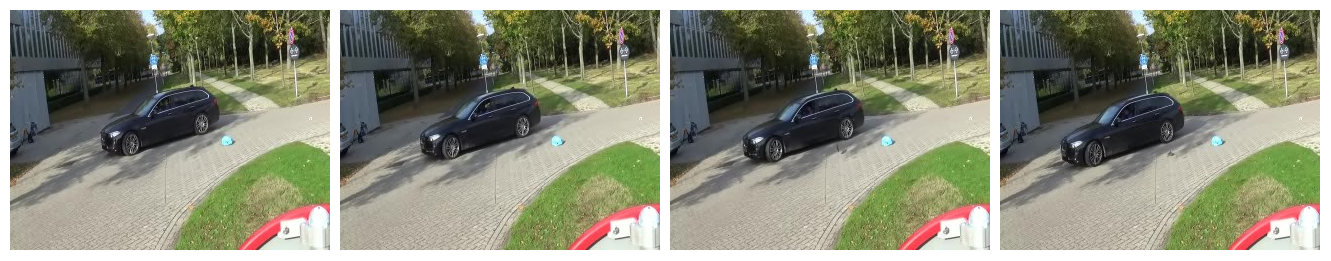}\end{tabular} \\ \hline
Pedestrian Dangerous & \begin{tabular}[c]{@{}c@{}}Object is thrown by a pedestrian\\ towards another traffic participant\end{tabular}                   &
\begin{tabular}[c]{c}\includegraphics[width=0.3\textwidth]{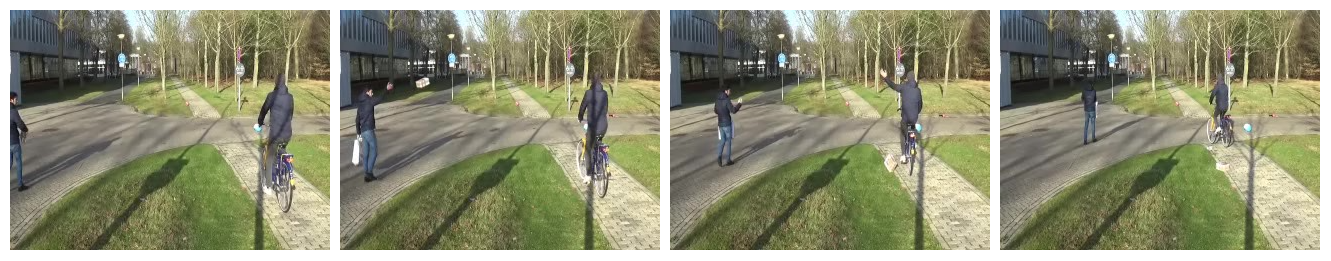}\end{tabular} \\ \hline
Pedestrian Safe      & \begin{tabular}[c]{@{}c@{}}Object is thrown by a pedestrian\\ onto the ground\end{tabular}                                       &
\begin{tabular}[c]{c}\includegraphics[width=0.3\textwidth]{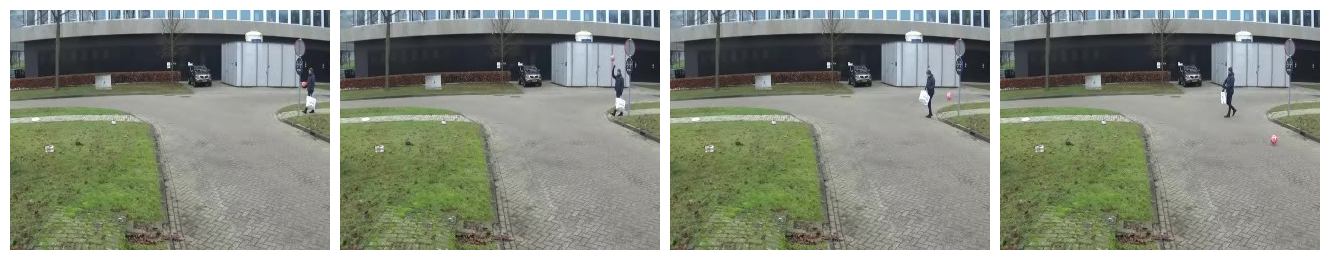}\end{tabular} \\ \hline
Normal               & \begin{tabular}[c]{@{}c@{}}No throwing actions occur but\\ some of the traffic participants do\\ occur in the video\end{tabular} &
\begin{tabular}[c]{c}\includegraphics[width=0.3\textwidth]{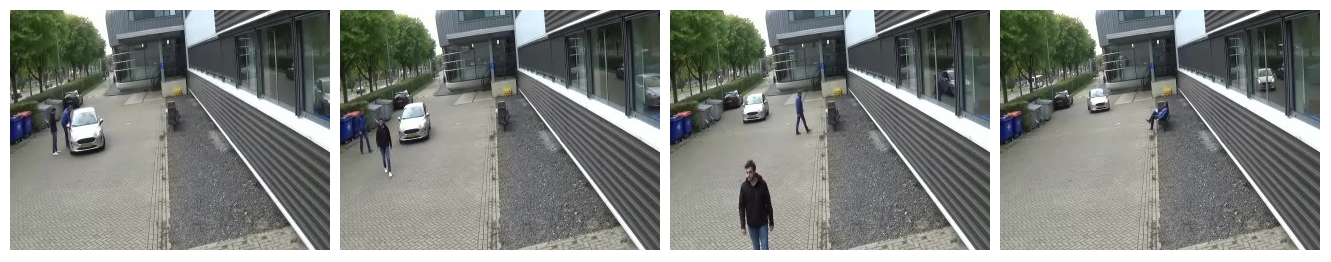}\end{tabular}\\
\hline
\multicolumn{1}{l}{}
\end{tabular}
\caption{Table 1. Definition and examples of each video category in the Throwing-Action dataset.}
\label{tab:Video_examples}
\end{table*}

\begin{table}
\centering
\begin{tabular}{l|cc|c}
\hline
                     & \multicolumn{3}{c}{Throwing-Action} \\
                     & Training & Testing  &  Total        \\ \hline \hline
Bicycle Dangerous    & 42    & 21    &  63      \\
Bicycle Safe         & 20    & 10    &  30      \\
Car Dangerous        & 38    & 19    &  57      \\
Car Safe             & 22    & 11    &  33      \\
Pedestrian Dangerous & 38    & 20    &  58      \\
Pedestrian Safe      & 20    & 10    &  30      \\
Normal               & 87    & 43    &  130     \\
\hline 
\multicolumn{1}{l}{}
\end{tabular}
\caption{Table 2. Distribution of the number of videos over the various classes for the Throwing-Action dataset.}
\label{tab:dataset_counts}
\end{table}

Currently, datasets are not available from literature to train a throwing anomaly detector. Therefore, we have generated a novel dataset, including the throwing anomalies in six different outdoor categories to address the anomaly use case from the Smart City project. The throwing anomalies are split into classes based on the acting traffic participant (car, bicycle, pedestrian) performing the throw, and whether or not the thrown object is directed towards another traffic participant. If the throw is directed at another traffic participant, it is called a 'dangerous' throw, otherwise a 'safe' throw. The generated dataset includes six throwing anomaly classes as described in Table~\ref{tab:Video_examples}. All videos have a resolution of 320$\times$240 pixels. One of the challenges in our dataset is that throwing anomaly videos can contain multiple throwing actions up to a maximum of ten. 

The objects used to create the throwing anomaly dataset are selected to have a large diversity in object shape, size, and color. Additionally, some objects maintain their shape during a throwing action, such as a football, while others become deformed and change their shape during the throwing movement, such as a sweater or plastic bag. Overall, the generated dataset consists of 130~normal videos without throwing anomalies and 271~anomalous videos divided over all six anomaly classes, titled as the \textit{"Throwing-Action"} dataset throughout the paper. Table~\ref{tab:dataset_counts} provides the number of videos of each anomaly class in the dataset.

\subsection{A. Annotation}
In the training set, each video is labeled as either normal or anomalous at the video sequence level, while each video is annotated at the frame level in the testing set. In other words, the start and end frames of each throwing action are provided for the test set. While the start frame of an anomaly action is defined as the first frame in which the object starts moving, the end frame of it is considered as the first frame where the object touches the ground or is occluded.

\subsection{B. Training and Testing Sets}
The generated "Throwing-Action" dataset is divided into a training and testing set. The training set consists of 87~normal videos and 180~anomalous videos, while the testing set contains the other 43~normal videos and 91~anomalous videos. Table~\ref{tab:dataset_counts} shows the number of videos for each class in both subsets.


\section{Methodology}
\label{Methodology_section}

\begin{figure*}[!ht]
    \centering
    \includegraphics[width=0.9\textwidth]{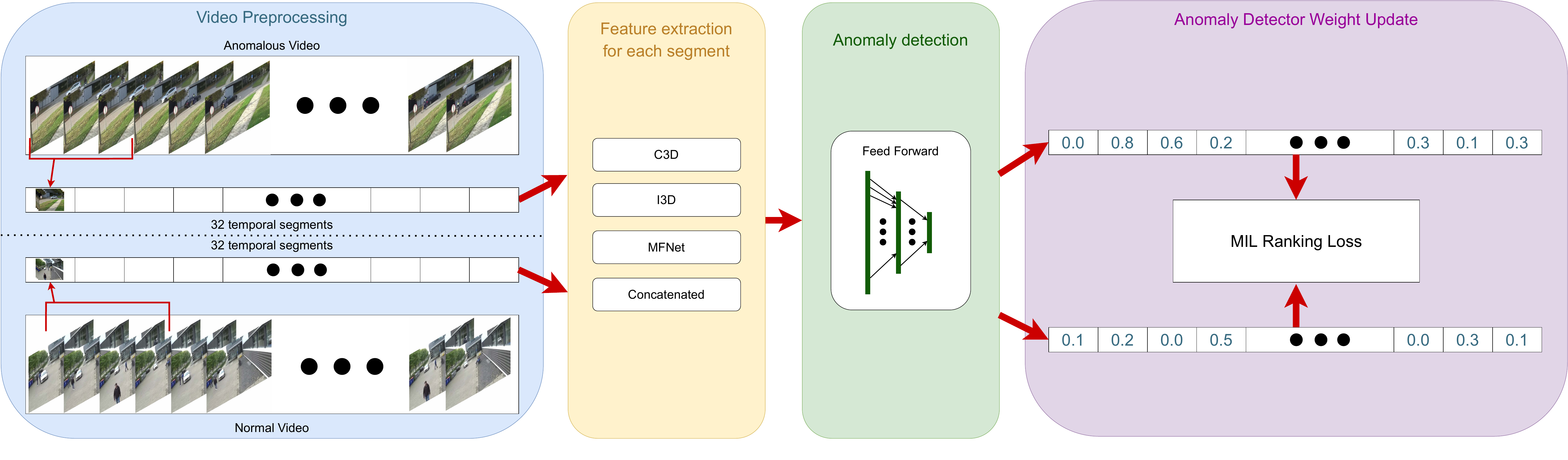}
    \caption{Flow diagram of the proposed methodology. Each video is split into 32 temporal segments. The videos are represented as a bag, and the segments as instances within this bag. Features are then extracted from each segment by the pre-trained C3D, I3D, or MFNet feature extraction networks. Next, these features are provided with an anomaly score by a fully connected neural network, resulting in one anomaly score for every instance in the bag. The network uses a multiple instance learning ranking loss for training.}
    \label{fig:flow_diagram_method}
\end{figure*}

\subsection{A. Baseline Method}
In this paper, we adopt an anomaly detection method inspired by~\cite{Sultani2018} as a baseline model, based upon the implementation of~\cite{anomaly18cvpr-pytorch}. Figure~\ref{fig:flow_diagram_method} depicts a diagram of the anomaly detection method. During the training process, input videos are split into 32~non-overlapping segments of equal length. These segments can be considered as instances in a bag, where the bag represents the video as a whole. To describe these segments, feature vectors are extracted from every 16-frame video clip within the instance, using a pre-trained feature extractor network. To represent the instance, we take the average of all 16-frame video clip features within that segment. Next, the instance-wise feature vectors are supplied to a feed-forward network, which aims to predict the likelihood that the input instance contains an anomaly, resulting in the representation of each instance in the bag by the predicted anomaly score. A multiple instance learning ranking loss ensures that one anomalous and one normal video are used for each training step. The notations $\mathcal{V}_a$ and $\mathcal{V}_n$ represent an anomalous and a normal video, respectively, while $\mathcal{A}_a^i$ and $\mathcal{A}_n^i$ represent the predicted anomaly score of the i-th segment of an anomalous and a normal video, respectively. The multiple-instance learning ranking loss function can then be expressed by:
\begin{equation}
    \label{eq:loss}
    \mathcal{L}_{\text{orig}}(\mathcal{V}_a,\mathcal{V}_n) = \max \ ( \ 0 \ , \ 1 - \underset{i \in \mathcal{V}_a}{\max}f(\mathcal{A}_a^i)+\underset{i \in \mathcal{V}_n}{\max}f(\mathcal{A}_n^i) \ ).
\end{equation}
This loss function forces the maximum anomaly score of the normal video towards zero, while simultaneously increasing the maximum anomaly score of the anomalous video towards unity. The assumption is made that the segment with the highest anomaly score is the true anomaly. 

Next to this multiple-instance learning ranking loss component, three more components are added to the loss in order to ensure temporal smoothness, sparsity and small model weights with scaling factor $\lambda_1$, $\lambda_2$ and $\lambda_3$, respectively. These additions result in the objective function given by 
\begin{equation}
    \label{eq:objective_function}
    \begin{split}
        \mathcal{L}_{\text{obj}} & = \mathcal{L}_{\text{orig}}(\mathcal{V}_a,\mathcal{V}_n) \ + \ \lambda_1 \sum_i^{(n-1)}(f(\mathcal{A}_a^i) - f(\mathcal{A}_a^{i+1}))^2 \\
        & + \ \lambda_2 \sum_i^nf(\mathcal{A}_a^i) \ + \  \lambda_3 \lVert \mathcal{W} \rVert_F \ ,
    \end{split}
\end{equation}
where $\mathcal{W}$ represents the model weights, and $n$ the number of segments into which a video is split.

\subsection{B. Variations on the Baseline Method}
We experiment with different feature extractors within our methodology and forms of our loss functions, which both are discussed briefly below. The paper that inspired our proposal applied the C3D network~\cite{Tran2015} only, whereas the further investigated feature extractors are the I3D network~\cite{Carreira2017QuoDataset}, MFNet network~\cite{Chen2018}, and features concatenated from the aforementioned extractors.

\textit{C3D network}: The C3D network is originally proposed in~\cite{Tran2015} and is one of the first applications of a three-dimensional convolutional neural network (3D-CNN) for supervised video action classification. We extract the output of the first fully connected layer as our features. In this work, the C3D feature extractor is pre-trained on the Sports1M dataset~\cite{Karpathy2014Large-scaleNetworks}. 

\textit{I3D network}: The I3D network~\cite{Carreira2017QuoDataset} is based on the Inception-V1~\cite{Ioffe2015BatchShift} detector. We take the output of the average-pooling layer as our computed features, resulting in a 1024-dimensional feature vector. Here, the I3D network is pre-trained on two different datasets, the Charades dataset~\cite{Sigurdsson2016HollywoodUnderstanding} and the Kinetics dataset~\cite{Carreira2017QuoDataset}. These feature extraction backbones are referred to as I3D-charades and I3D-kinetics, respectively, throughout the paper.

\textit{MFNet network}: The MFNet network~\cite{Chen2018} is used to extract features from the input videos. In literature, this network is not yet applied for the purpose of anomaly detection in videos. The MFNet network achieves slightly better performance than I3D on action-recognition datasets, while at the same time, it requires up to ten times fewer computations according to~\cite{Chen2018}. The MFNet feature extraction backbone is pre-trained on the UCF-101 dataset~\cite{DBLP:journals/corr/abs-1212-0402}. The MFNet features are represented by 6,144-dimensional vectors.

\textit{Concatenated features}: Finally, we concatenate all the extracted features from the C3D, I3D-charades, I3D-kinetics and MFNet networks together, of which we construct a 12,288-dimensional feature vector. The anomaly detection model has access to a wider range of information, thereby improving anomaly detection performance with the help of these concatenated features. The downside of this approach is the increase in computational cost to execute all feature extraction networks.

In our method, the loss is based only on the maximum obtained anomaly score of any segment in a video, both for normal and anomalous videos. For an anomalous video, this makes sense because it could be possible that such a video only contains one anomalous segment. However, all segments should ideally have an anomaly score of zero for normal videos. Hence, we experiment with a loss function that takes the mean predicted anomaly score into account, as opposed to only the maximum. This loss is from now on referred to as the mean normal loss and described formally below. This change allows the model to learn from all normal segments at every training iteration, so that the mean normal loss is specified by
\begin{equation}
    \label{eq:loss_adapted}
    \mathcal{L}_{\text{mean-nl}}(\mathcal{V}_a,\mathcal{V}_n) = \max \ ( \ 0 \ , \ 1 - \underset{i \in \mathcal{V}_a}{\max}f(\mathcal{A}_a^i)+\underset{i \in \mathcal{V}_n}{\text{mean}}f(\mathcal{A}_n^i) \ ).
\end{equation}
Finally, we apply different optimizers in order to compare their effects on the model performance. The baseline paper~\cite{Sultani2018}, uses the Adadelta optimizer, while we also apply the Adam optimizer.


\section{Experiments}
\label{Experiment_section}

In this section, we perform experiments with our anomaly detection method described in the \nameref{Methodology_section}~section on the UCF-Crime dataset and the generated Throwing-Action dataset to improve the anomaly detection performance. The primary metric used to compare anomaly detection performance between experiments is the area under the Receiver Operating Characteristic (ROC) curve. This metric is commonly used to evaluate the performance of anomaly detection methods and it gives insight into the achieved ratio between the true-positive rate and false-positive rate for the full range of anomaly thresholds. 

\begin{figure}[t]
    \centering
    \includegraphics[width=0.45\textwidth]{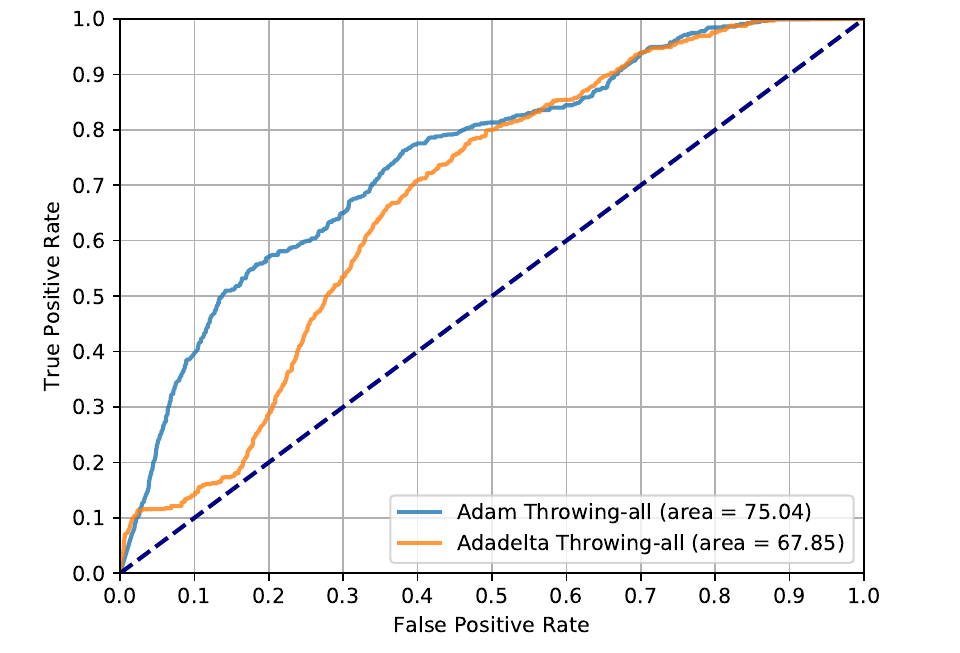}
    \caption{Optimal ROC curves obtained on the testing set of the Throwing-Action dataset when the anomaly detection model is trained with the Adadelta and Adam optimizers.}
    \label{fig:throwing_ROC}
\end{figure}
    
\subsection{A. Comparison of Adadelta and Adam Optimizers}

Our first contribution is determining the appropriate optimizer for the training phase of all experimentation models. We train an anomaly detection model on the Throwing-Action training dataset for 100,000~iterations, using the Adadelta optimizer with a learning rate of~0.01, similarly to~\cite{Sultani2018} and using the Adam optimizer with a learning rate of~0.0005. Figure~\ref{fig:throwing_ROC} depicts the optimal ROC curves for each model.

\begin{figure}
    \centering
    \includegraphics[width=0.45\textwidth]{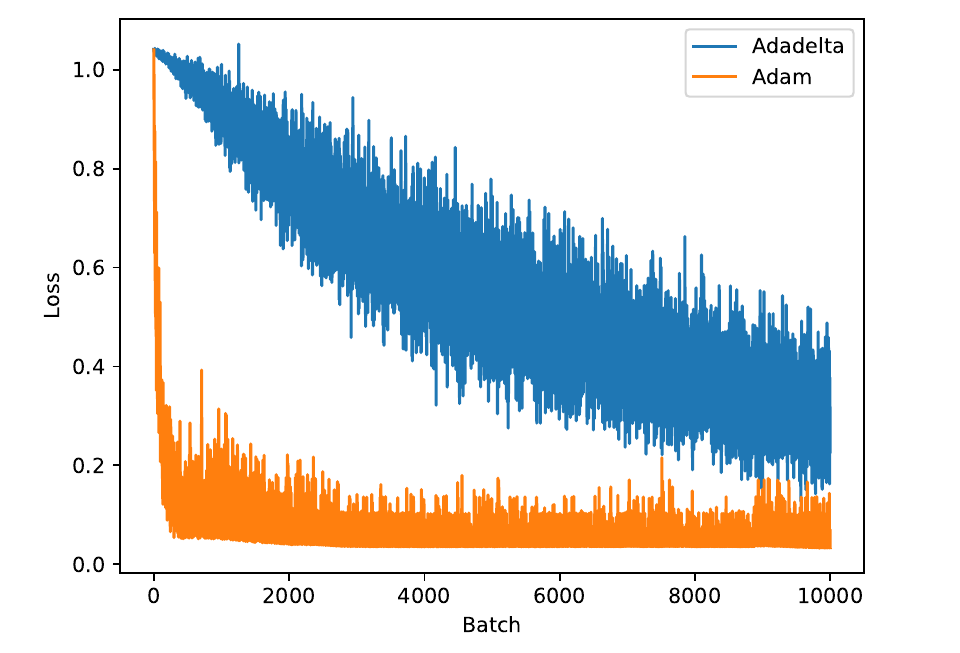}
    \caption{Training loss curves of the anomaly detection model on the Throwing-Action dataset for Adadelta and Adam optimizers, showing the mean loss of every batch during training.}
    \label{fig:loss_ada_vs_adam}
\end{figure}

As shown in Figure~\ref{fig:throwing_ROC}, Adam outperforms Adadelta by a significant margin. This difference indicates that the Adadelta training process has not reached an optimum after 100,000~iterations. Figure~\ref{fig:loss_ada_vs_adam} shows the batch loss during training for both the Adadelta and Adam optimizers. From this graph, it can be observed that the learning process is significantly slower for the Adadelta optimizer. Therefore, the Adam optimizer is found more suitable to train anomaly detection models. Since our anomaly detection model trained with the Adam optimizer performs better and has a faster learning process, all models are from now on trained using the Adam optimizer with a learning rate of~0.0005.

\subsection{B. Video Augmentation}

Next, several approaches for data augmentation are explored to improve performance. Augmentation methods are applied to the videos of the Throwing-Action dataset to diversify the training data, which helps the model in generalizing to new cases. Furthermore, test-time augmentation (TTA) is explored for augmentation as well.

\begin{table}
\centering
\begin{tabular}{l|c|c|c}
\hline
             & \begin{tabular}[c]{@{}c@{}}No\\ Augment.\end{tabular} & Augment.  & \begin{tabular}[c]{@{}c@{}}Test Time\\ Augment.\end{tabular} \\ \hline \hline
C3D          & 75.04            & 76.23          & 75.55                                                             \\
I3D-charades & 78.71            & 80.83          & 83.15                                                            \\
I3D-kinetics & 80.41            & 82.84          & 84.32                                                             \\
MFNet        & \textbf{86.10}   & 83.63          & 83.89                                                    \\
Concatenated & 85.29            & \textbf{85.58} & \textbf{85.52} \\  
\hline
\multicolumn{1}{l}{}
\end{tabular}
\caption{Table 3. Maximum area under the ROC curve achieved on the Throwing-Action dataset for all feature extraction networks, when applying different augmentation methods (best scores in bold).}
\label{tab:Throwing-Action_augmentation}
\end{table}

\begin{figure}[t]
    \centering
    \includegraphics[width=0.45\textwidth]{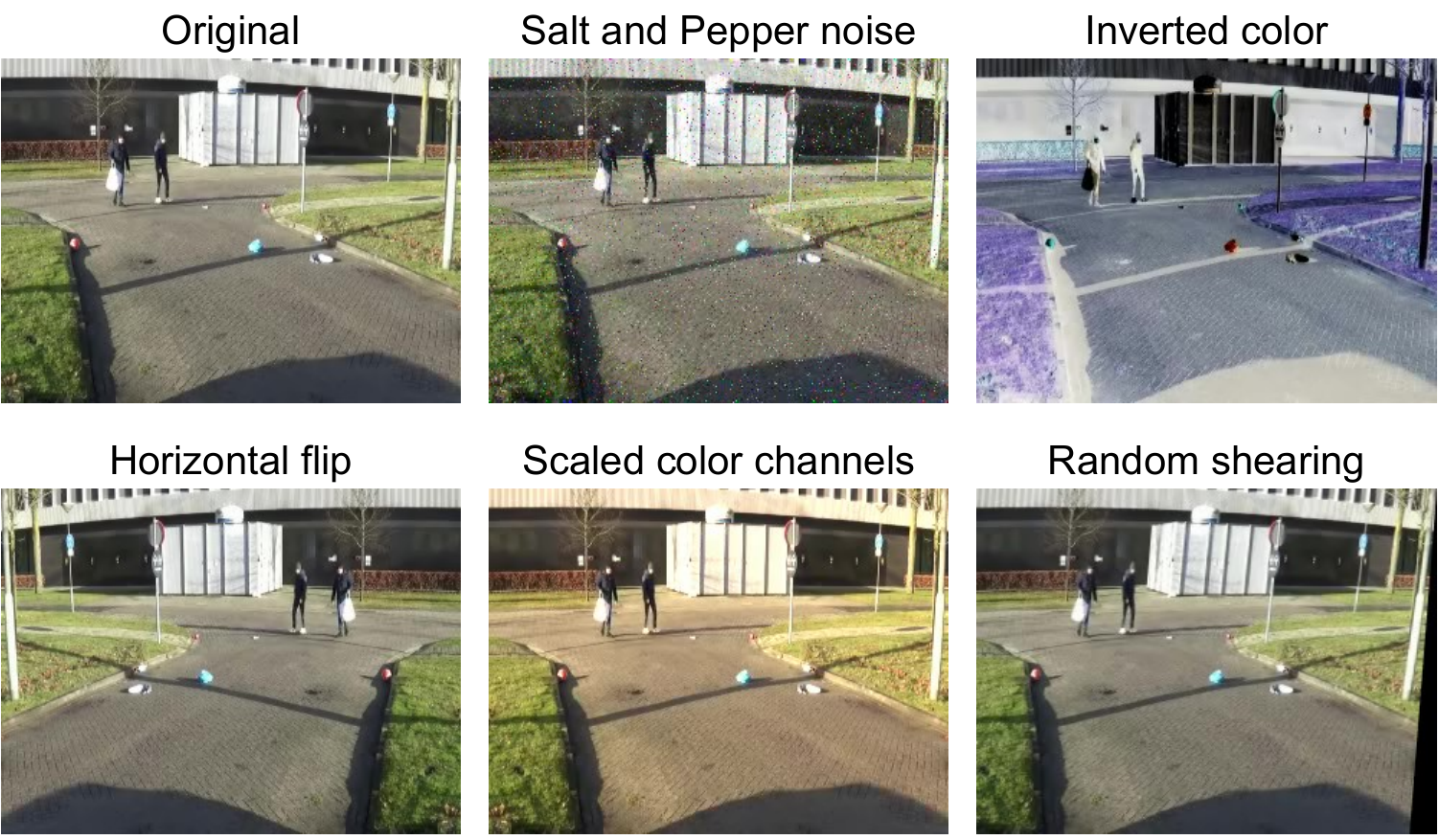}
    \caption{Visual examples of each type of image augmentation applied to the Throwing-Action dataset. From left to right and top to bottom: original, (no augmentation), salt and pepper noise, inverted color, horizontal flip, independently scaled color channels and random shearing.}
    \label{fig:augmentations}
\end{figure}

\subsubsection{1) Training set augmentation}
We have applied five augmentation methods to the original videos. The augmentation methods used are salt-and-pepper noise, where in total 3\% of all sub-pixels are affected; inverting colors; horizontal flipping; independently scaling the color channels by a factor between 0.8 and 1.2; and finally, applying shear transformations. Visual examples of each augmentation output are illustrated in Figure~\ref{fig:augmentations}.

An anomaly detection model is trained on both the original Throwing-Action dataset and the augmented Throwing-Action dataset for each feature extraction network. Table~\ref{tab:Throwing-Action_augmentation} shows the largest area under the ROC curve obtained on the test set of the Throwing-Action dataset for all feature extractors. From this table, we can conjugate that the choice of the feature extraction has a significant impact on the quality of the resulting model. The C3D feature extraction network performs worst, while MFNet and the concatenated features perform best. Training with the concatenated features does not improve the performance of any individual feature extractor. The MFNet features contain all necessary information for detecting the anomalies, while the other feature extractors effectively only manage to extract a subset of this information. This is visible because MFNet network obtains the highest score, while the other ones and even the concatenated network score lower. Table~\ref{tab:Throwing-Action_augmentation} indicates that training on augmented data improves the obtained results for all feature extractors, except for MFNet. We obtain the best results for the MFNet feature extractor without augmentation while for concatenated features augmentation is needed. However, due to the significantly increased computational cost of the concatenated feature extractor compared to the MFNet feature extractor, the final model uses MFNet without data augmentation.

\subsubsection{2) Test time augmentation (TTA)}
Another technique that can improve the performance of the anomaly detection model is test time augmentation (TTA). The videos in the test set are augmented using the same augmentation methods as used for the training set, while predictions of the anomaly detection model are averaged over all versions of the same video. Table~\ref{tab:Throwing-Action_augmentation} shows the maximum area under the ROC curves when training the anomaly detection model on the augmented Throwing-Action dataset and testing with TTA. TTA further improves the results for both versions of the I3D feature extractors, but does not manage to improve results for other feature extractors. Furthermore, TTA requires augmented test data which makes the model more expensive at test time.

In conclusion, since the optimal area under the ROC curve is obtained by a model trained without any augmentation and training with augmentation is more computationally expensive, we have decided to leave out data augmentation in the remainder of this paper.

\subsection{C. Experiments on UCF-Crime+Throwing Dataset}

In addition to anomaly detection performance on the newly generated Throwing-Action dataset, we are interested in combining the new Throwing-Action dataset with the publicly available UCF-Crime~\cite{Sultani2018} anomaly dataset. Therefore, we first evaluate the performance achieved on the UCF-Crime dataset for all feature extractors. 

Table~\ref{tab:Extended} shows the best maximum area under the ROC curve on the UCF-Crime dataset for all feature extraction backbones. It indicates that the performance for the C3D feature extraction is significantly lower than other feature extractors. Another observation from Table~\ref{tab:Extended} is that concatenated features outperform any individual feature extraction backbone. For the UCF-Crime dataset, the different feature extractors provide different sets of information about the video, and concatenating these allows the detector model to consider more information. Furthermore, comparing the performance on the UCF-Crime dataset with the performance on the Throwing-Action dataset shown in Table~\ref{tab:Throwing-Action_augmentation} reveals that for each feature extraction network the performance on the UCF-Crime dataset is lower, leading to the conclusion that the UCF-Crime dataset is overall a more difficult dataset.

\begin{table}
\centering
\begin{tabular}{l|c|c|c|c}
\hline
             & Exp. 1          & Exp. 2          & Exp. 3         & Exp. 4         \\ \hline \hline
C3D          & 70.30           & 69.80           & 67.68          & 70.66          \\
I3D-charades & 78.36           & 77.95           & 79.80          & 77.64          \\
I3D-kinetics & 78.47           & 78.34           & 80.25          & 78.01          \\
MFNet        & 76.76           & 72.12           & 75.19          & 71.50          \\
Concatenated & \textbf{79.84}  & 79.16           & \textbf{81.08} & 78.86          \\ \hline
\begin{tabular}[l]{@{}l@{}}Concatenated\\ Mean normal loss\end{tabular}
             & -               & \textbf{80.13}  & 78.59          & \textbf{80.31} \\
\multicolumn{1}{l}{}
\end{tabular}
\caption{Table 4. Maximum area under the ROC curve achieved for different experiments. Exp 1: model trained and evaluated on UCF-Crime dataset. Exp 2, 3 and 4: model trained on UCF-Crime+Throwing dataset, evaluated on UCF-Crime+Throwing, Throwing-Action and UCF-Crime datasets, respectively.}
\label{tab:Extended}
\end{table}

Next, we combine the UCF-Crime dataset with our Throwing-Action dataset. This new dataset is from now on referred to as UCF-Crime+Throwing. In order to compare the performance, an anomaly detection model is trained on the training subset of the UCF-Crime+Throwing dataset for each feature extraction backbone and evaluated on the UCF-Crime+Throwing, the Throwing-Action, and the UCF-Crime testing sets. Table~\ref{tab:Extended} shows the maximum area under the ROC curve obtained on each test set for every feature extraction backbone. The results indicate that the performance on the UCF-Crime+Throwing dataset is generally similar to the performance on only the UCF-Crime dataset. The results show that it is possible to add throwing anomaly detection capabilities at only a small cost in the detection performance of other anomalies. Therefore, it is viable to integrate throwing anomaly detection into general anomaly detection systems.

The final experiment is concerned with the proposed mean normal loss function. Since this loss enables the model to learn from all normal segments at every training iteration, changing the loss function allows the model to better recognize normal sections of a video, which decreases the false positive rate and increases the area under the ROC curve. Table~\ref{tab:Extended} summarizes the performance for this modified loss function when using concatenated features and training on the UCF-Crime+Throwing dataset. This model achieves the highest overall performance on the UCF-Crime+Throwing dataset, but at the cost of decreased performance on the Throwing-Action testing set.

\subsection{D. Qualitative results}

\begin{figure*}
    \centering
    \includegraphics[width=0.9\textwidth]{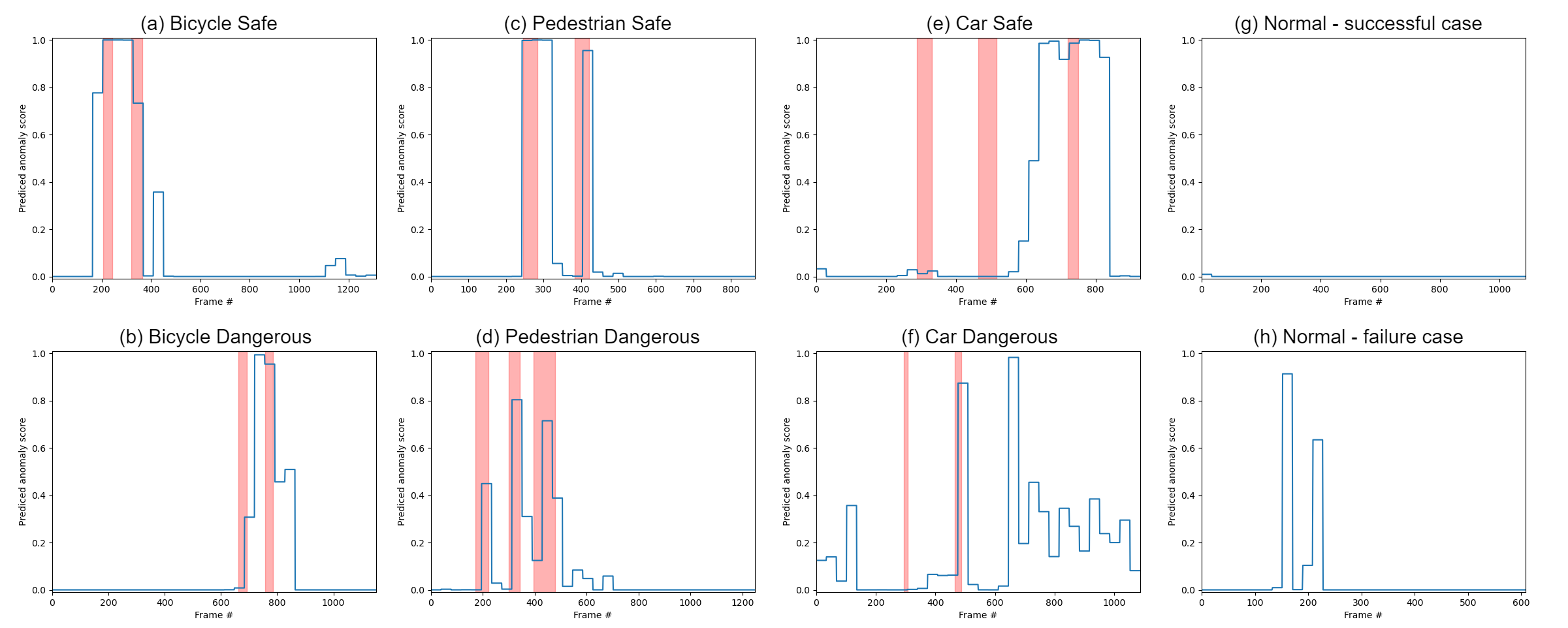}
    \caption{Qualitative results of the anomaly detection method on testing videos from the Throwing-Action dataset. Red regions show ground-truth anomalous regions, while the blue curve depicts the predicted anomaly score as a function of the frame number. From left to right and from top to bottom, the test videos cover anomaly classes: (a) bicycle safe, (c) pedestrian safe, (e) car safe, (g) successful case normal, (b) bicycle dangerous, (d) pedestrian dangerous, (f) car dangerous. Subfigure (h) shows a failure case for the normal video class.}
    \label{fig:visualizations}
\end{figure*}

This section presents qualitative results of the proposed anomaly detection method. Figure~\ref{fig:visualizations} shows the output of the feed-forward network as a function of time, expressed by the frame number of several testing videos of the Throwing-Action dataset. This means that the $y$-axis shows the likelihood that a segment is anomalous, according to our anomaly detection method, and the $x$-axis shows the progression number of processed video frames. Furthermore, the ground-truth anomalous regions within each video are given as red colored areas. In Figure~\ref{fig:visualizations}~(a)-(d), it can be observed that the performance is quite good for pedestrian and bicycle-related anomalies. However, several segments have a high predicted anomaly score, while being outside of the ground-truth anomalous areas. This problem occurs most often between two distinct anomalies or just after one anomaly. These segments contain relatively large person-arm motions after completing a throw, which may seem similar to a throwing action. 

In Figure~\ref{fig:visualizations}~(e)(f) show predictions for car-related throwing anomalies, which are more difficult to identify for our anomaly detection method. Anomalous sections sometimes obtain a predicted anomaly score of zero. This is caused by the fact that throwing anomalies from cars are more challenging to recognize, as there could be no person-arm motions. Furthermore, the second peak with a predicted anomaly score above~0.5 in subfigure~(f) corresponds to the moment that the car suddenly accelerates in the video. An explanation for this high score is that the normal training videos do not contain sufficient video material of accelerating cars, making this acceleration appear anomalous. 

Finally, subfigures~(g)(h) show predicted anomaly scores of two different normal videos. As can be noticed, the example in (g) is close to the ideal performance for a normal video, with a low predicted anomaly score in every segment. However, subfigure (h) indicates two false alarms, which are most likely caused by the presence of a fire truck in the footage of this normal video, because they rarely occur in any normal training video.


\begin{table}[ht]
\centering
\begin{tabular}{l|c|c}
\hline
                 & Original loss & Proposed loss \\ \hline \hline
False alarm rate & 0.5667       & \textbf{0.4696} \\       
\hline \multicolumn{1}{l}{}
\end{tabular}
\caption{Table 5. False alarm rate in percentage of normal video segments where the predicted anomaly score is more than 0.5 on the combined UCF-Crime+Throwing dataset.}
\label{tab:False_alarm_rate}
\end{table}

Overall, the experiments show that the best results for the combined UCF-Crime+Throwing dataset are achieved when training the anomaly detection model with the Adam optimizer on concatenated features, and with the proposed mean normal loss function using the mean anomaly score of normal videos. The UCF-Crime+Throwing dataset contains a wider range of anomalies, compared to the proposed Throwing-Action dataset. This variation explains why the concatenated features increase the performance for the combined dataset, while they do not provide an improvement for the Throwing-Action dataset. From the qualitative analysis, we conclude that our throwing anomaly detection model works well for pedestrian and bicycle-related anomalies. However, performance for car-type anomalies can be further improved by involving more data and situations appearing with cars.


\section{False Alarm Analysis}
\label{Analysis_section}

This section separately analyzes the performance of the proposed anomaly detection method in terms of false alarm rate because it is an important indicator for the correct focus on safety.

\begin{figure}[t!]
    \centering
    \includegraphics[width=0.48\textwidth]{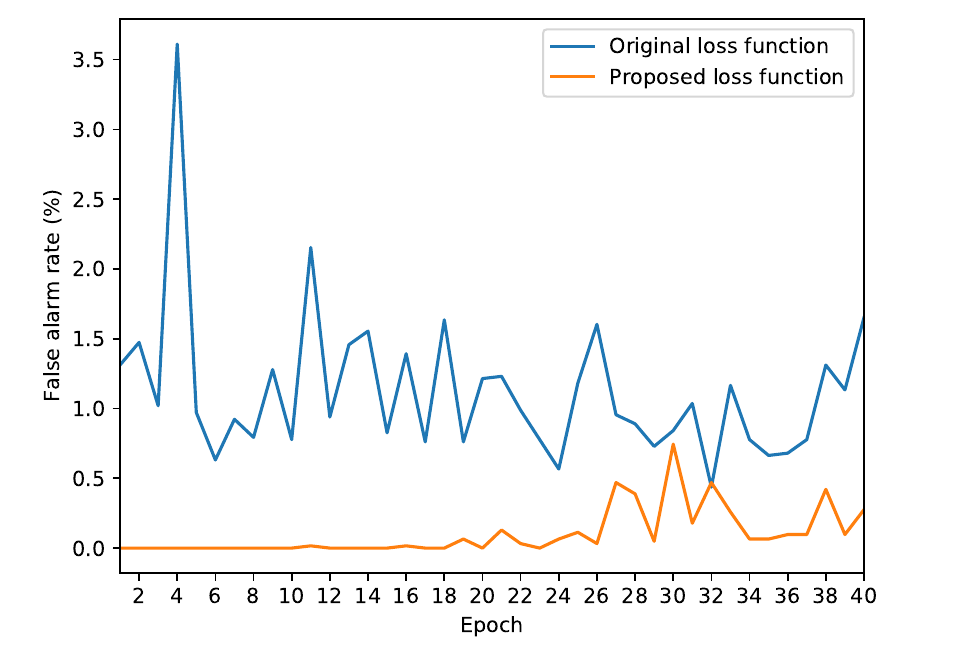}
    \caption{False alarm rate on normal testing data for anomaly detection models trained using the original loss function and the mean normal loss function. Both models are trained on the combined UCF-Crime+Throwing dataset and false alarms are categorized as a predicted anomaly score of more than 0.5 on a normal video segment.}
    \label{fig:False_alarm_rate}
\end{figure}

A false alarm rate of even a few percent can render an anomaly detection algorithm useless, since by definition most of the data encountered by an anomaly detection method is normal. Therefore, the false alarm rate is an important metric for industrial applications. We compare the false alarm rates on the normal testing data of the combined UCF-Crime+Throwing dataset for models trained on concatenated features with the original loss function and the mean normal loss function, as shown in Figure~\ref{fig:False_alarm_rate}. A false alarm is defined as a predicted anomaly score above~0.5 for a segment of a normal video. This figure shows that changing the loss function appears to positively affect the false alarm rate, reducing it significantly. With the proposed loss function, the model learns to recognize normal data faster, which results in an initial false alarm rate of zero. As the training progresses, the model learns to recognize anomalies, which sometimes introduce erroneous predictions.


\section{Conclusion}
\label{Conclusion_section}


We have introduced a novel dataset consisting of normal behaviors and throwing actions performed by car drivers, pedestrians, and bicyclists, to support the development of throwing anomaly detection algorithms. Additionally, we have compared the performance of different feature extraction networks, which are C3D, I3D-charades, I3D-kinetics and MFNet, within our anomaly detection method on the Throwing-Action dataset, the UCF-Crime dataset, and the combined dataset. Furthermore, we have found that using the Adam optimizer for training anomaly detection algorithms leads to improved results and shorter training times over the Adadelta optimizer. Finally, using the proposed mean normal loss function, we have significantly reduced the false alarm rate of the anomaly detection method, thereby improving the area under the ROC curve. Based on the experimental results, we have successfully improved the performance on the combined dataset to an area under the ROC curve of~80.13 by means of the proposed mean normal loss function, Adam optimizer and concatenated features.



\section{Acknowledgments} 

This work is supported by the European ITEA project SMART on intelligent traffic flow systems and Efficient Deep Learning (EDL) RMR project.



\small
\bibliographystyle{IEEEtranS}


\begin{biography}

Ivo Kersten obtained a B.S. degree in electrical engineering from the Eindhoven University of Technology in 2020, and a M.S. degree in electrical engineering with a specialization in artificial intelligence engineering systems from the same university in 2022. For his thesis work he conducted research on the detection of object throwing behavior in surveillance videos. 

Erkut Akdag received the B.S. and M.S. degrees in electrical and electronics engineering from Middle East Technical University, Ankara, Turkey, in 2012 and 2015, respectively. He is currently working toward the Ph.D. degree at the Video Coding and Architectures Group, Eindhoven University of Technology, Eindhoven, The Netherlands. His research interests include critical vehicle detection, anomaly detection and computer vision for surveillance. 

Egor Bondarev obtained his PhD degree in the Computer Science Department at TU/e, in research on performance predictions of real-time component-based systems on multiprocessor architectures. He is an Assistant Professor at the Video Coding and Architectures group, TU/e, focusing on sensor fusion, smart surveillance and 3D reconstruction. He has written and co-authored over 50 publications on real-time computer vision and image/3D processing algorithms. He is involved in large international surveillance projects like APPS and PS-CRIMSON.

Peter H.N. de With is Full Professor of the Video Coding and Architectures group in the Department of Electrical Engineering at Eindhoven University of Technology. He worked at various companies and was active as senior system architect, VP video technology, and business consultant. He is an IEEE Fellow and member of the Royal Holland Society of Academic Sciences and Humanities, has (co-)authored over 600 papers on video coding, analysis, architectures, and 3D processing and has received multiple papers awards. He has served as a program committee member of various IEEE conferences and holds some 30 patents.
\end{biography}

\end{document}